\documentclass[10pt,twocolumn,letterpaper]{article}

\usepackage{iccv}              

\definecolor{iccvblue}{rgb}{0.21,0.49,0.74}
\usepackage[pagebackref,breaklinks,colorlinks,allcolors=iccvblue]{hyperref}

\usepackage{adjustbox,multirow,makecell,tabularx}

\def\paperID{3677} 
\def\confName{ICCV}
\def\confYear{2025}

\title{TokenFLEX: Unified VLM Training for Flexible Visual Tokens Inference}

\author{
    Junshan Hu$^{1}$\thanks{Equal contribution.} \quad
    Jialiang Mao$^{1*}$\quad
    Zhikang Liu$^{1}$\thanks{Corresponding author.} \quad
    Zhongpu Xia$^{1}$ \quad
    Peng Jia$^{1}$\quad
    Xianpeng Lang$^{1}$ \\
    $^{1}$Li Auto\\
}

\begin{document}
\maketitle
\pagestyle{plain}

\begin{abstract}

Conventional Vision-Language Models (VLMs) typically utilize a fixed number of vision tokens, regardless of task complexity. 
This one-size-fits-all strategy introduces notable inefficiencies: using excessive tokens leads to unnecessary computational overhead in simpler tasks, whereas insufficient tokens compromise fine-grained visual comprehension in more complex contexts. 
To overcome these limitations, we present TokenFLEX, an innovative and adaptable vision-language framework that encodes images into a variable number of tokens for efficient integration with a Large Language Model (LLM). 
Our approach is underpinned by two pivotal innovations. 
Firstly, we present a novel training paradigm that enhances performance across varying numbers of vision tokens by stochastically modulating token counts during training.
Secondly, we design a lightweight vision token projector incorporating an adaptive pooling layer and SwiGLU, allowing for flexible downsampling of vision tokens and adaptive selection of features tailored to specific token counts. 
Comprehensive experiments reveal that TokenFLEX consistently outperforms its fixed-token counterparts, achieving notable performance gains across various token counts—enhancements of 1.6 \%, 1.0\%, and 0.4\% with 64, 144, and 256 tokens, respectively—averaged over eight vision-language benchmarks. 
These results underscore TokenFLEX's remarkable flexibility while maintaining high-performance vision-language understanding.

\end{abstract}

\section{Introduction}
Vision-language models (VLMs)~\cite{liu2024visual,liu2024improved,liu2024llava,chen2023internvl,chen2024howfar,bai2023qwen-vl,wang2024qwen2vl,lu2024deepseek,wu2024deepseek,yao2024minicpm} have become foundational in multi-modal artificial intelligence, facilitating advancements in tasks such as visual question answering, image captioning, and cross-modal retrieval. Commercial models like GPT-4o~\cite{hurst2024gpt4o} and Claude~\cite{anthropic2024claude} consistently extend the capabilities of VLMs. Presently, the open-source community is rapidly advancing, achieving performance on par with, or even exceeding, that of proprietary models by improving model architectures, data strategy, and training recipes~\cite{liu2025ola,li2024llavaOV,chen2025internvl25,bai2025qwen25vl}. However, a fundamental limitation persists: existing VLMs enforce a \textit{fixed vision token count} for all tasks, creating a rigid trade-off between task-specific requirements and model performance. Simple tasks (e.g., coarse category classification) are burdened with redundant tokens, while complex tasks (e.g., fine-grained scene analysis) suffer from insufficient visual detail due to token scarcity. This \textit{one-size-fits-all} design forces models to compromise accuracy for tasks requiring precise visual understanding or waste computational resources on simpler scenarios.

\begin{figure}
    \centering
    \includegraphics[width=1.0\linewidth]{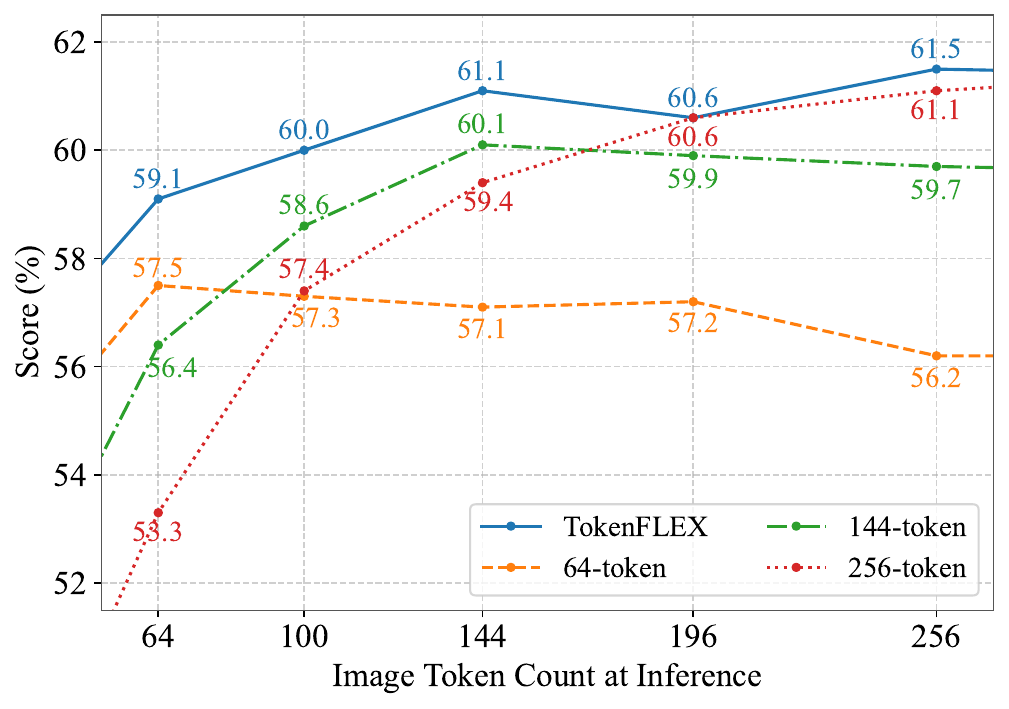}
    \caption{Comparison of TokenFLEX with methods using fixed vision token counts. TokenFLEX is trained with a stochastic dynamic image token count, while other methods are trained with fixed token counts of 64, 144, and 256, respectively. The performance is measured as the average score across 8 multi-modal benchmarks on OpenCompass~\cite{2023opencompass}. TokenFLEX consistently outperforms the fixed-token methods, particularly when using fewer tokens, such as 64.}
    \label{fig:TokenFLEX_intro}
\end{figure}

Current frontier VLMs face significant limitations in adjusting the number of vision tokens during inference, primarily due to two key challenges.
The first challenge stems from the fact that these models are trained with a fixed number of vision tokens. For instance, LLaVA~\cite{liu2024llava,li2024llavaOV} consistently uses 729 tokens per image. Altering this number during inference can lead to out-of-distribution (OOD) issues, resulting in a notable decline in performance. As illustrated in Figure~\ref{fig:TokenFLEX_intro}, a model trained with a fixed 256 tokens exhibits a 4.2\% drop in the OpenCompass~\cite{2023opencompass} metric when using 64 tokens during inference, compared to a model explicitly trained with 64 tokens. 
The second challenge lies in the design of the projector, which is typically built for a fixed number of vision tokens and does not support arbitrary modifications. For example, LLaVA-OneVision~\cite{li2024llavaOV} employs an MLP to map vision features, while PixelShuffle in InternVL~\cite{chen2023internvl,chen2025internvl25} only supports a fixed 4x downsampling. Similarly, MiniCPM-V~\cite{yao2024minicpm} relies on 64 predefined learnable queries. Although some projectors~\cite{yao2024deco,li2024tokenpacker,liu2024oryx} theoretically support changing the visual token numbers during inference, their effectiveness still requires further exploration.

In this paper, we propose TokenFLEX, a novel vision-language framework that liberates VLMs from fixed visual token constraints. Our approach enables flexible adjustment of visual token quantities based on task requirements while enhancing the projector architecture to support dynamic token configurations. Our key contributions are as follows:
\textit{1) Dynamic Token Mechanism.} To address the fixed token constraint in training, we develop a stochastic training method that randomly select the number of vision token from a predefined set for each sample. This forces the model to learn robust cross-model alignment consistency across varying token numbers, effectively mitigating the OOD performance degradation during inference. This training paradigm supports flexible inference with arbitrary token counts while maintaining accuracy comparable to fixed-token baselines.
\textit{2) Lightweight Token-Adaptive Projector.} To address the weakness of previous projectors, we design a new lightweight token-adaptive projector that incorporates an adaptive average pooling layer and SwiGLU~\cite{shazeer2020glu}. The adaptive average pooling layer enables flexible modification of the number of vision tokens, while SwiGLU leverages its gate mechanisms to dynamically assign weights to visual features, prioritizing salient information under varying token configurations while suppressing redundant details. This adaptive weighting ensures critical visual semantics are preserved even when token counts are reduced.

Extensive experiments across a broad range of vision-language benchmarks demonstrate that TokenFLEX consistently outperforms various fixed-token baselines. 
Notably, TokenFLEX shows improvements of 1.6\%, 1.0\%, and 0.4\% with 64, 144, 256 tokens, respectively, averaged over eight benchmarks when compared to fixed-token baselines.
Additionally, the proposed flexible token-length encoding mechanism further reduces training costs, in our experiments, it decreased visual token usage by up to 28\% and shortened training time by 13\%, enabling more efficient model training without sacrificing performance. 
\section{Related Work} \label{sec:related_work}
\begin{figure*}[!ht]
    \centering
    \includegraphics[width=1.0\linewidth]{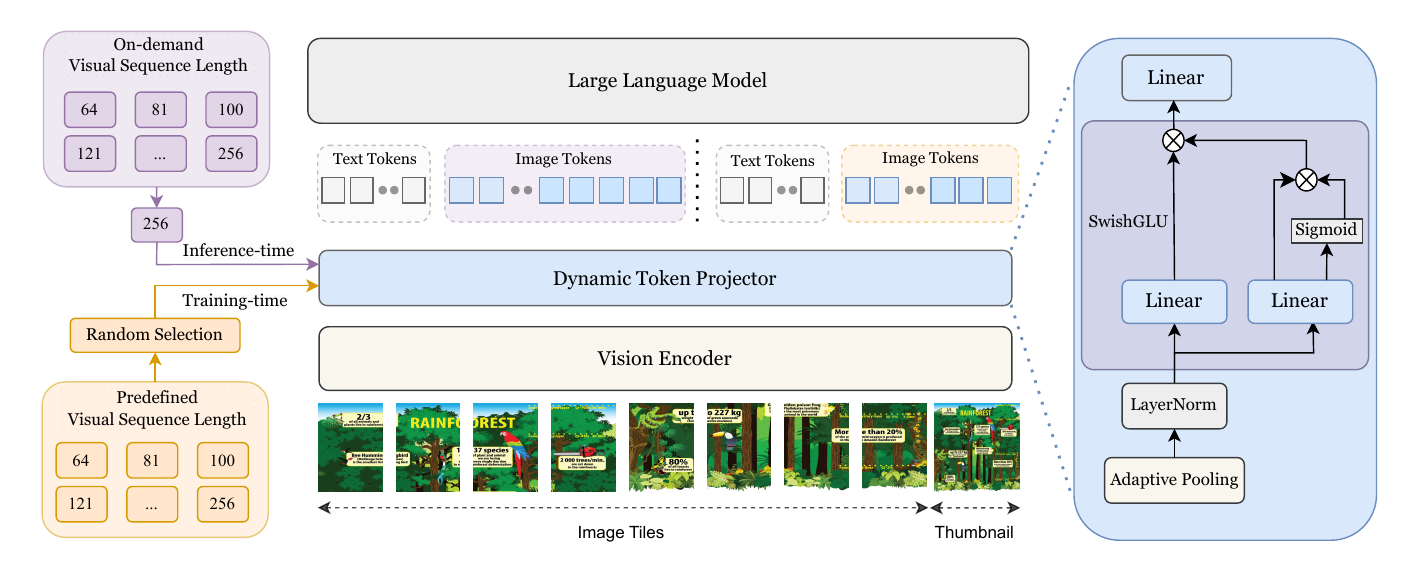}
    \caption{Overall architecture. TokenFLEX combines a Visiual Encoder and a Large Language Model connected by a dynamic token projector. (Left) During the training phase, each sample randomly selects an image token length; during inference, the image token length can be chosen on-demand based on task complexity and computational budget. (Right) Our light-weight projector acts as an adaptive filter, allowing it to selectively emphasize important image tokens. }
    \label{fig:tokenFLEX_arch}
\end{figure*}

\subsection{Large Vision-Language Models (VLMs)}
Recent advances in Vision-Language Models (VLMs) \cite{liu2024visual,liu2024improved,liu2024llava,chen2023internvl,chen2024howfar,bai2023qwen-vl,wang2024qwen2vl,lu2024deepseek,wu2024deepseek,yao2024minicpm,liu2023llava,li2023blip2,chen2025internvl25,liu2024oryx} have demonstrated remarkable cross-modal understanding capabilities. 
Pioneering works like Flamingo \cite{alayrac2022flamingo} established the paradigm of fusing frozen vision encoders (e.g., CLIP \cite{radford2021CLIP}) with large language models (LLMs) through cross-modal projectors. 
This architecture has become the standard for modern VLMs: visual features extracted by vision encoders are projected into \textit{vision tokens} via lightweight adapters (e.g., linear layers \cite{liu2023llava} or Q-Former modules \cite{li2023blip2}), which are then concatenated with text tokens as input to LLMs. 
Recent advancements, exemplified by LLaVA family \cite{liu2023llava,li2024llavaOV}, Qwen2-VL \cite{wang2024qwen2vl}, and InternVL2.5 \cite{chen2025internvl25}, have significantly improved multimodal understanding capabilities through scaled-up training datasets and optimized training recipes.

However, most existing methods suffer from a critical limitation: they adopt \textit{fixed vision token allocation} (e.g. 256 tokens per image \cite{chen2025internvl25}) regardless of input complexity or computational constraints. This rigid paradigm leads to suboptimal efficiency-accuracy tradeoffs. 
While some methods \cite{wang2024qwen2vl,liu2024oryx} using native resolution ViT \cite{dehghani2023navit} could support dynamic token allocation, they require training proprietary ViTs that limit model extensibility (e.g., inability to integrate diverse vision encoders for improved perception \cite{shi2024eagle,kar2024brave}). Although Oryx-MLLM \cite{liu2024oryx} proposes allocating different token numbers for different modalities, its cannot specify arbitrary token counts. In contrast, our method supports variable vision token allocation during inference, enabling flexible selection of vision token numbers according to task complexity. This novel paradigm maintains competitive performance while providing unprecedented flexibility.

\subsection{Projector Design for Cross-Modal Alignment}
The projector plays a critical role in VLMs by aligning visual and linguistic representations and reducing computational cost through token compression. Current methods can be categorized into two main paradigms.
The first category employs linear projector or multilayer perceptrons (MLPs) for alignment. Examples like LLaVA \cite{liu2023llava} and InternVL2 \cite{chen2023internvl} map vision encoder outputs directly to the language model's embedding space. 
To address token redundancy, Pixel-shuffle \cite{chen2023internvl}, adaptive average pooling \cite{yao2024deco}, and convolutional layers \cite{chu2023mobilevlm} have been proposed to compress visual tokens efficiently. 
The second paradigm utilizes learnable query mechanisms with cross-attention layers to extract important visual features, such as Q-Former \cite{li2023blip2} and Resampler \cite{bai2023qwen-vl}. 
Recent innovations like TokenPacker \cite{li2024tokenpacker} and Oryx-MLLM \cite{liu2024oryx} initialize queries with downsampled ViT features, subsequently refining them through local cross-attention layers.

Despite these advancements, most projectors are designed with a fixed number of visual tokens, limiting their flexibility to accommodate varying token numbers. For example, both Pixel-shuffle \cite{chen2023internvl} and Resampler \cite{bai2023qwen-vl} require a predetermined downsample ratio. Although pooling-based methods \cite{yao2024deco,li2024tokenpacker,liu2024oryx} theoretically support dynamic tokens, their effectiveness with dynamic tokens remains underexplored.
We systematically investigates projector architectures that enable robust dynamic token processing. Through comprehensive experimentation, we identify optimal design choices that maintain performance stability across varying token budgets while preserving cross-modal alignment capabilities.
\section{Method}

In this section, we present the design details of TokenFLEX, which consists of a vision encoder, a lightweight projector, and a large language model, as shown in Figure~\ref{fig:tokenFLEX_arch}. The framework offers two core advantages: \textit{flexible visual token adjustment} during inference and \textit{enhanced performance}, achieved through two key designs: 1) a \textit{dynamic token mechanism} (Section~\ref{sec:dynamic_token}), and 2) a \textit{lightweight token-adaptive projector} (Section~\ref{sec:projector}).

\subsection{Overview of TokenFLEX}

TokenFLEX adopts a standard Vision-Language Model (VLM) framework based on the Vision Transformer (ViT)-Projector-Large Language Model (LLM) architecture. For a given image $\mathbf{I}_{\text{img}} \in \mathbb{R}^{H \times W \times 3}$, the ViT extracts a visual representation $\mathbf{F}_{\text{vis}} \in \mathbb{R}^{L \times C_{\text{v}}}$, where $L = H' \cdot W' = (H/p \cdot W /p)$ is the number of visual features. Here $p$ represents the patch size of ViT, and $C_{\text{v}}$ denotes to the dimension of vision feature. Subsequently, the proposed lightweight token-adaptive projector $\mathcal{P}(\cdot)$ compresses these $L$ visual features into $N$ vision tokens and aligns them with the textual embedding space of the LLM:
\begin{equation}
    \mathbf{T}_{\text{vis}} = \mathcal{P}(\mathbf{F}_{\text{vis}}, N)
\end{equation}

where $N$ is flexibly specified based on demand. To preserve spatial consistency, $N$ is constrained to be a perfect square, ensuring alignment with the ViT's output grid structure. The vision tokens are concatenated with text embeddings $\mathbf{T}_{\text{text}}$ and fed into the LLM for autoregressive generation. The output sequence $\mathbf{R} = [r_1, r_2, \ldots, r_M]$ is formalized as:

\begin{equation}
P(\mathbf{R}|\mathbf{T}_{\text{vis}}, \mathbf{T}_{\text{text}}) = \prod_{m=1}^{M} P(r_m | \mathbf{R}_{<m}, \mathbf{T}_{\text{vis}}, \mathbf{T}_{\text{text}}).
\end{equation}

\subsection{Dynamic Token Mechanism} \label{sec:dynamic_token}

Traditional VLMs employ a fixed number of visual tokens during both training and inference, which restricts the flexible adjustment of vision token counts during inference according to task complexity or computational budget. This rigid approach results in suboptimal trade-offs between efficiency and accuracy. To overcome the limitation of fixed visual token counts in traditional VLMs, we propose a \textit{dynamic token mechanism}, allowing for the flexible adjustment of $N$ at test time. 

However, directly varying $N$ in fixed-token models causes performance degradation. As shown in Figure~\ref{fig:TokenFLEX_intro}, a model trained with 256 visual tokens drops by 4.2\% in performance when tested with 64 vision tokens, compared to the model specifically trained with 64 vision tokens. This indicates overfitting to fixed token counts, leading to out-of-distribution (OOD) issues when $N$ changes.

To address these issues, we propose a stochastic dynamic token training method, which randomly select the visual token count $N$ from predefined set \( \Phi = \{n_1, n_2, ..., n_t\} \) for each training loop. To balance the performance across different vision token counts, we adjust the proportion of each vision token count in the training set based on probabilities \( \Psi = \{p_1, p_2, ..., p_t\} \), where each probability \(p_i\) corresponds to the vision token count \(n_i\) with \(\sum_{i=1}^{t} p_i = 1\). 
The stochastic selection of \(N\) can be expressed as follows:
\begin{equation}
    P(N = n_i) = p_i \quad \text{for } i = 1, 2, \ldots, t
    \label{eq:random_token_count}
\end{equation}

To enhance efficiency, batches are homogenized to contain a single $N$ value. This approach offers three advantages: 1) it alleviates OOD issues by enabling the LLM to adapt to diverse token counts; 2) the diversity of token counts acts as an implicit data augmentation, allowing learning across different token counts to mutually benefit and thus improve model performance; 3) reduced $N$ during training minimizes computational overhead. As illustrated in Table~\ref{tab:ablation_dynamic_token}, our experimental results demonstrate that the stochastic dynamic token training strategy can achieve or even surpass the performance of fixed token training.

\subsection{Lightweight Token-Adaptive Projector} \label{sec:projector}
The projector is critical for aligning visual features with textual embeddings while reducing computational costs. Existing methods like MLP with Pixel-Shuffle \cite{chen2023internvl} and Resampler \cite{bai2023qwen-vl,yao2024minicpm} lack flexibility in token count adjustment, hindering dynamic token inference. Inspired by previous works~\cite{yao2024deco,li2024tokenpacker,liu2024oryx}, we designed a new projector architecture that supports dynamic token mechanism, as shown in Figure~\ref{fig:tokenFLEX_arch}. This module incorporates a Swish-Gated Linear Unit (SwiGLU)~\cite{shazeer2020glu}, which can adaptively select useful features, facilitating more flexible adaptation to varying numbers of visual tokens.

Our projector first employs an adaptive average pooling layer to restructure the visual feature maps into a user-specified token count $N$, achieved by partitioning the feature grid into an \(n \times n\) spatial layout (where $N = n^{2}$), where the features within each grid are grouped to form \(\mathbf{F}_{\text{vis}}^g \in \mathbb{R}^{N \times H'/n \times W'/n \times C_\text{v}}\).
Each group features is then pooled into a single feature, resulting in \(\mathbf{F}_{\text{vis}}^p \in \mathbb{R}^{N \times 1 \times C_{\text{v}}}\). The pooled features \(\mathbf{F}_{\text{vis}}^p \) are first normalized via a LayerNorm denoted as  \(\mathbf{F}_{\text{vis}}^n \) and then processed through a Swish-Gated Linear Unit (SwiGLU)~\cite{shazeer2020glu}, which dynamically gates and fuses feature channels. A final linear projection layer $\textbf{W}_3$ maps the transformed features into the LLM's textual embedding space, generating semantically aligned vision tokens \(\mathbf{T}_{\text{vis}}\) that match the textual embedding space of the LLM. The projector is formulated as:
\begin{align}
  \mathbf{F}_{\text{vis}}^p &= \text{AdaptiveAvgPool} \left( \mathbf{F}_{\text{vis}}, N \right) \\
  \mathbf{F}_{\text{vis}}^n &= \text{LayerNorm} \left( \mathbf{F}_{\text{vis}}^p \right) \\
  \mathbf{T}_{\text{vis}} &= \mathbf{W}_3 \left( (\mathbf{W}_1 \mathbf{F}_{\text{vis}}^n) \odot \sigma(\mathbf{W}_2 \mathbf{F}_{\text{vis}}^n) \right)
\end{align}
where $\sigma(x)$ is the sigmoid function, $\textbf{W}_2$ and $\textbf{W}_3$ are two learnable weight matrices acts as linear layers.

This design allows seamless adaptation to arbitrary token counts specified at inference time, as validated in our ablation study (Section~\ref{sec:ablation}) against dynamic token projectors like TokenPacker~\cite{li2024tokenpacker} and Ola~\cite{liu2025ola}.

\begin{table*}[t]
    \centering
       \begin{tabular}{llccc}
            \toprule
            & & \textbf{Stage-1} & \textbf{Stage-1.5} & \textbf{Stage-2} \\
            \midrule
            \multirow{2}{*}{\textbf{Vision}} & Resolution & $384$ & \multicolumn{2}{c}{$384 \times \{(i,j) \mid i, j \in \mathbb{Z}^+, i \times j \leq 12\}$} \\
            & Tokens ($\Phi$) & $\{64, 144, 256\}$ & \multicolumn{2}{c}{$(i \times j + 1) \times \{64, 144, 256\}$} \\ 
            \midrule
            \multirow{2}{*}{\textbf{Data}} & Dataset & Caption & Caption \& OCR & High-Quality Data \\
            & \#Samples & 2M & 1.5M & 2M \\
            \midrule
            \multirow{2}{*}{\textbf{Model}} & Trainable & Projector &  \makecell{Vision encoder,\\Projector}  & Full model \\
            & \#Parameters & 20.4M & 0.5B & 8.1B \\
            \midrule
            \multirow{4}{*}{\textbf{Training}} & Batch Size & 512 & 256 & 256 \\
            & Learning Rate & $1 \times 10^{-3}$ & $2 \times 10^{-5}$ & $2 \times 10^{-5}$ \\
            & Max Length & 4096 & 8192 & 8192 \\
            & Probabilities ($\Psi$) & $\{0.2, 0.3, 0.5\}$ & $\{0.2, 0.3, 0.5\}$ & $\{0.2, 0.3, 0.5\}$ \\
            \bottomrule
        \end{tabular}
    \caption{Configurations for the three stage training process of TokenFLEX in our experiments.}
    \label{tab:training_settings}
\end{table*}
\section{Experiments}

\subsection{Implementation Details}
In the implementation of TokenFLEX, we employ SigLIP-400M/14-384px~\cite{zhai2023siglip} as our visual encoder and Qwen2.5-7B-Instruct~\cite{qwen2.5} as our language decoder to leverage the strengths of both models in their respective domains. Given that the default resolution of the visual encoder is 384, we splice images into tiles to support higher resolutions, as demonstrated in previous works~\cite{li2024llavaOV,chen2023internvl}. We set a maximum of 12 tiles, and a thumbnail is also employed to provide a global overview.  Drawing inspiration from recent studies~\cite{li2024llavaOV,chen2025internvl25,shi2024eagle}, the training pipeline for TokenFLEX is structured into three stages, aimed at enhancing the model's visual perception and multimodal capabilities, as detailed in Table~\ref{tab:training_settings}.  

\textbf{Stage 1: Modality Alignment.} In the initial stage, only the projector is learnable with the aim of effectively aligning the visual features into the word embedding space of LLMs. Due to the stochastic dynamic token training, we utilize a relatively larger dataset of 2 million samples compared to previous work~\cite{li2024llavaOV}. The data consists of image captions randomly selected from 10M Caption data of Infinity-MM~\cite{gu2024infinitymm}, with images primarily sourced from LAION-2B~\cite{schuhmann2022laion} and CapsFusion-120M~\cite{yu2024capsfusion}. The learning rate is set to $1e^{-3}$.

\textbf{Stage 1.5: Vision Enhancement.} 
In this stage, the focus shifts to enhancing the vision encoder’s capacity to extract comprehensive visual features. Both the vision encoder and projector are trainable during this phase. The data primarily consists of captions and OCR from sources such as Docmatix~\cite{laurençon2024building}, ShareGPT4V~\cite{chen2024sharegpt4v}, Cambrian~\cite{tong2024cambrian}, and Ureader Caption~\cite{ye2023ureader}, among others. These datasets are collected by LLaVA-OneVision~\cite{li2024llavaOV} and Infinity-MM~\cite{gu2024infinitymm}. The learning rate is set at $2e^{-5}$, and the maximum length of LLM is increased to 8192 to accommodate high resolution inputs.

\textbf{Stage 2: Vision Instruction Tuning.} In the final stage, the full model is trained on 2 million high quality data points to enhance its conversational capabilities. The data, sourced entirely from LLaVA-OneVision~\cite{li2024llavaOV}, encompasses a diverse range of tasks,, such as general QA, math/reasoning, general OCR and language, among others.

\begin{table*}[ht]
    \centering
    \begin{tabular}{c|cc|cccccccc}
    \toprule
    Method                         & \makecell{\#Training-\\Token}             & \makecell{\#Inference-\\Token} & \rotatebox{45}{MMStar} & \rotatebox{45}{OCRB} & \rotatebox{45}{AI2D} & \rotatebox{45}{HallB} & \rotatebox{45}{MMB\textsubscript{1.1}} & \rotatebox{45}{MMVet} & \rotatebox{45}{MathVista} & \rotatebox{45}{MMMU} \\ \midrule
    \multirow{9}{*}{\rotatebox{0}{Baseline}} & \multirow{3}{*}{64}         & 64        & 53.0   & 64.1     & 78.4 & 39.6      & 76.0         & 48.0  & 52.5            & 48.1      \\
     &                      & 144 & 53.7 & 62.0 & 77.6 & 39.8 & 75.8 & 48.5 & 52.4 & 47.3 \\
     &                      & 256 & 53.2 & 59.5 & 77.3 & 40.1 & 75.1 & 45.4 & 50.9 & 48.4 \\ \cmidrule(lr){2-11}
     & \multirow{3}{*}{144} & 64  & 53.9 & 54.6 & 78.1 & 40.6 & 75.9 & 45.9 & 52.0 & 50.3 \\
     &                      & 144 & 55.1 & 67.4 & 80.0 & 42.6 & 77.1 & \textbf{51.9} & 55.5 & 51.3 \\
     &                      & 256 & 55.2 & 65.5 & 79.7 & 42.7 & 77.9 & 49.9 & 55.8 & 51.4 \\ \cmidrule(lr){2-11}
     & \multirow{3}{*}{256} & 64  & 50.0 & 45.5 & 74.8 & 39.4 & 75.1 & 40.7 & 49.7 & \textbf{51.3} \\
     &                      & 144 & 54.9 & 61.8 & 79.1 & \textbf{43.1} & 78.1 & 50.7 & 55.9 & \textbf{52.1} \\
     &                      & 256 & 55.3 & 69.9 & 80.2 & 43.9 & 78.9 & \textbf{51.8} & \textbf{56.9} & \textbf{52.2} \\ \cmidrule(lr){1-11}
    \multirow{3}{*}{\rotatebox{0}{TokenFLEX}}     & \multirow{3}{*}{64,144,256} & 64        & \textbf{54.4}   & \textbf{64.5}     & \textbf{79.4} & \textbf{42.2}      & \textbf{79.0}         & \textbf{49.0}  & \textbf{54.6}            & 49.6      \\
     &                      & 144 & \textbf{56.5} & \textbf{69.8} & \textbf{80.8} & 43.0 & \textbf{79.5} & 51.6 & \textbf{57.5} & 50.0 \\
     &                      & 256 & \textbf{57.3} & \textbf{71.4} & \textbf{81.0} & \textbf{44.4} & \textbf{79.9} & 51.0 & 56.5 & 50.6 \\
     \bottomrule
    \end{tabular}
    \caption{TokenFLEX performance on different benchmarks. We conduct experiments with different vision token counts during training and inference. Comparing with models using fixed single vision tokens, TokenFLEX achieves best results in most benchmarks.}
    \label{tab:compare_with_baseline}
\end{table*}

\subsection{Evaluation Benchmarks}
We conducted a comprehensive evaluation of tokeFLEX's performance across eight benchmarks used in the OpenCompass ranking~\cite{2023opencompass}. These benchmarks include MMBench~\cite{liu2023mmbench} and MMStar~\cite{mmstar} for assessing general abilities, MMMU~\cite{yue2024mmmu} for STEM skills, HallusionBench~\cite{liu2023hallusionbench} for hallucination testing, MathVista~\cite{lu2023mathvista} for mathematical competencies, AI2D~\cite{Kembhavi2016ai2d} for chart comprehension, OCRBench~\cite{liu2023ocrbench} for OCR capabilities, and MMVet~\cite{yu2023mmvet} for subjective evaluation. All results were evaluated using the Vlmevalkit~\cite{duan2024vlmevalkit}.

\subsection{Main Results}

To validate the effectiveness of the proposed TokenFLEX, we designed a fixed-token baseline, which uses the same ViT and LLM as TokenFLEX. The differences between the baseline and TokenFLEX are twofold. First, the baseline's projector employs a MLP with GELU activation, as adopted by previous works~\cite{chen2025internvl25,li2024llavaOV}, whereas TokenFLEX uses the proposed Token-Adaptive projector. Second, the baseline is trained using a fixed vision token count, while TokenFLEX employs the proposed stochastic dynamic token training method. We train the baseline on 64, 144, and 256 tokens, respectively, ensuring that all other training configurations remain consistent with TokenFLEX, as described in Table~\ref{tab:training_settings}. Figure~\ref{fig:TokenFLEX_intro} illustrates the average scores on OpenCompass, demonstrating that TokenFLEX consistently outperforms the fixed-token baselines, especially with fewer tokens, such as 64. Table~\ref{tab:compare_with_baseline} presents detailed results across eight benchmarks. TokenFLEX achieves superior performance in most benchmarks. Notably, increasing the number of inference tokens, yields significant performance improvements in benchmarks such as MMStar and OCRBench, while some benchmarks like MMBench and MMVet exhibit limited enhancement. Therefore, TokenFLEX can effectively balance performance and efficiency by flexibly adjusting the number of vision tokens during inference, a capability unattainable with fixed-token models. 

To compare with state-of-the-art models, we expand the Stage-2 training dataset to 7 million samples using Infinity-MM~\cite{gu2024infinitymm}. We then evaluate TokenFLEX on the OpenCompass. Table~\ref{tab:compare_with_sota} presents the results, showing that TokenFLEX achieves competitive performance across most benchmarks. Notably, TokenFLEX with 64 tokens scores 61.8\% on OpenCompass, comparable to LLaVA-OneVision, which uses 729 tokens. These results demonstrate that TokenFLEX not only supports flexible vision tokens inference but also achieves competitive performance against state-of-the-art models.

Table \ref{tab:training_time} presents a comparison of training efficiency performed on 64 A100 GPUs during Stage 2 of the training process. 
Compared with fixed 256-token training, TokenFLEX, with  $\Phi = \{64, 144, 256\}$ tokens, reduces the number of visual tokens from 7.8B to 5.6B, marking a 28\% decrease. It also shortens the training time from 15.0 hours to 13.0 hours, signifying a 13\% reduction.

\begin{table*}[]
    \centering
    \begin{tabular}{l|cc|cccccc}
        \toprule
        Model                      & Size               & \#Token & MMB\textsubscript{1.1} & MMMU & OCRBench & AI2D & HallB & \makecell{Open-\\Compass} \\ \midrule
        \multicolumn{9}{l}{\textit{Close-source Models}}                                                                               \\ \midrule
        GPT-4o-0513~\cite{openai2023gpt4o}      & -     & -       & 82.2     & 69.2 & 736    & 84.6    & 55.0    &  69.9    \\
        GPT-4V~\cite{openai2023gpt4v}           & -     & -      & 79.8      & 61.7 & 656    & 78.6    & 43.9    &  63.5    \\
        Gemini-1.5-Pro~\cite{team2024gemini}    &  -    & -       & -        & 62.2 & 754    & -       & -       &  64.4    \\ \midrule
        \multicolumn{9}{l}{\textit{Publicly Available Models}}                                                                         \\ \midrule
        LLaVA-OneVision~\cite{li2024llavaOV}  & 8B      &  729    & 80.9    & 46.8 & 697    & 82.8     & 47.5    &  61.2    \\
        MiniCPM-V2.6~\cite{yao2024minicpm}   & 8B        &  64     & 78.0    & 49.8 & 852    & 82.1     & 48.1    &  65.2    \\
        InternVL2.5~\cite{chen2025internvl25}  & 8B     &  256    & 82.5    & 56.2 & 821    & 84.6     & 49.0    &  68.1    \\
        Cambrian-1~\cite{tong2024cambrian} &  8B          & 576        & 68.2    & 41.8 & 614   & 74.6     &  30.6    & 52.9     \\
        Deepseek-VL~\cite{lu2024deepseek} &  7.3B              & 576     & 70.7    & 38.3 & 435    & 65.3     & 34.5     &  46.2    \\
        LLaVA-NeXT~\cite{liu2024llavanext} &  8B                & 576     & 69.8    & 43.1 & 531    & 72.8     & 33.1    & 49.7  \\
        VILA-1.5~\cite{lin2023vila} &  8B                & 196     & 57.9    & 37.4 & 438   & 58.8     & 35.3    & 44.0     \\ \midrule
        \multirow{3}{*}{TokenFLEX} & \multirow{3}{*}{8B}& 64      & 79.6    & 49.1 & 734   & 80.2     & 49.8    & 61.8     \\
                                   &                    & 144     & 80.0    & 49.2 & 771   & 81.5     & 51.2    & 63.9     \\
                                   &                    & 256     & 80.5    & 49.9 & 783   & 82.0     & 50.5    & 64.6     \\ \bottomrule
    \end{tabular}
    \caption{Performance comparison of tokenFLEX and stat-of-the-art models. In this experiment, we extend the training data of Stage-2 to 7 million samples to enhance the model's generalization ability. TokenFLEX achieves competitive results across most benchmarks.}
   \label{tab:compare_with_sota}
\end{table*}

\begin{table}[ht]
    \centering
    \begin{tabular}{c|cc}
    \toprule
    \#Training Tokens & \#Vision Tokens & Training Time (h) \\ \midrule
    64              & 2.0B            & 8.2           \\
    144             & 4.4B            & 10.8          \\
    256             & 7.8B            & 15.0           \\
    64,144,256      & 5.6B            & 13.0            \\
    \bottomrule
    \end{tabular}
    \caption{Comparison of training efficiency between TokenFLEX and fixed-token training in Stage 2. Testing is conducted on 64 A100 GPUs. TokenFLEX reduces the number of vision tokens by 28\% and decreases training time by 13.3\% relative to the fixed 256-token training.}
    \label{tab:training_time}
\end{table}

\begin{table}[]
    \centering
    \begin{tabularx}{\columnwidth}{c|*{5}{>{\centering\arraybackslash}X}}
    \toprule
    \multirow{2}{*}{\makecell{\#Training\\Token}} & \multicolumn{5}{c}{\#Inference Token}      \\ \cmidrule(lr){2-6} 
                            & 64    & 100   & 144   & 196   & 256   \\ \midrule
    64                        & 57.5 & 57.4 & 57.1 & 57.2 & 56.2 \\
    144                       & 56.4 & 58.6 & 60.1 & 60.0 & 59.8 \\
    256                       & 53.3 & 57.4 & 59.5 & 60.6 & 61.1 \\ \midrule
    64,144,256                & 58.7 & 59.3 & 60.0 & 60.1 & 60.7 \\ \bottomrule
    \end{tabularx}
    \caption{Ablation of dynamic token mechanism. Compared to fixed token training, dynamic token training achieved competitive results with 144, 196, and 256 tokens, and achieved the best results with 64 and 100 tokens.}
    \label{tab:ablation_dynamic_token}
\end{table}

\begin{figure*}
    \centering
    \includegraphics[width=\textwidth]{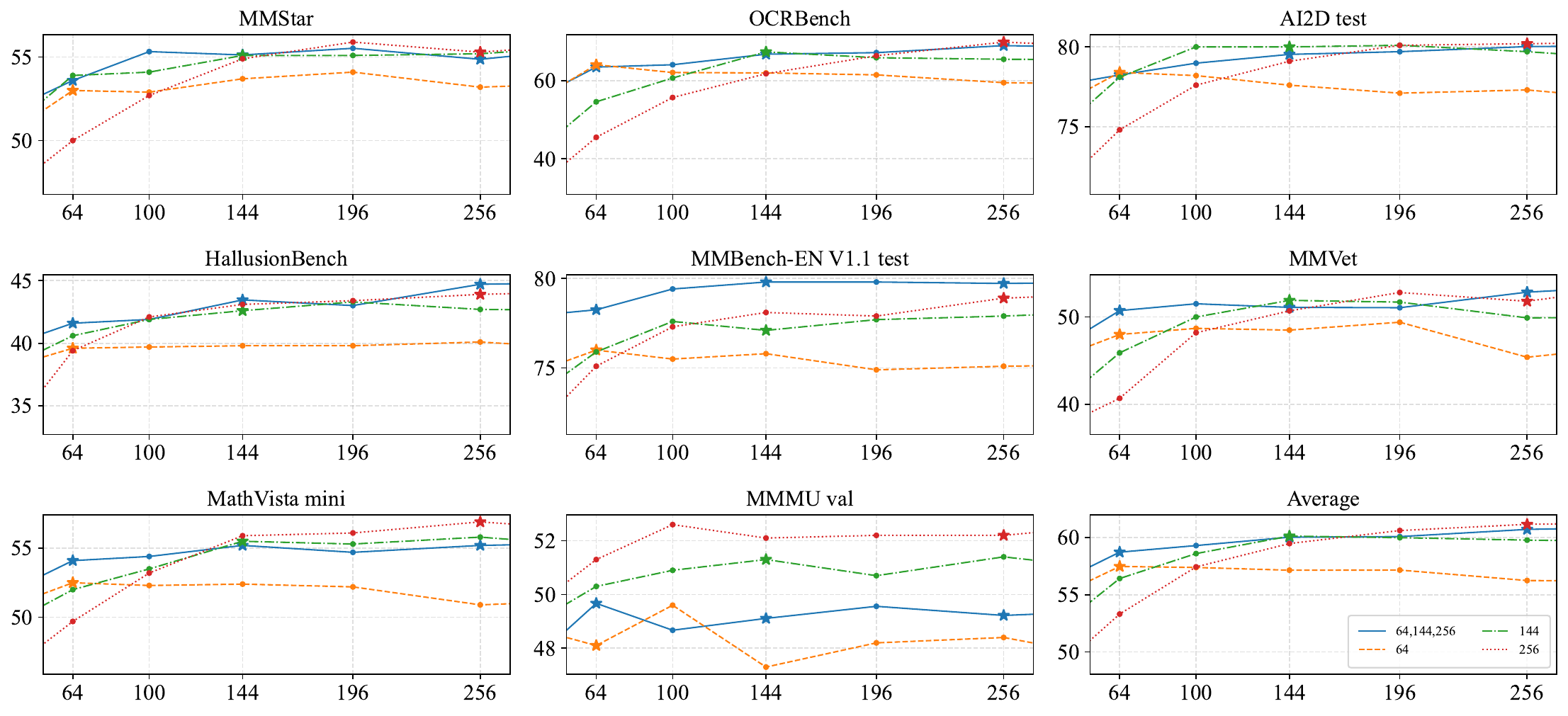}
    \caption{Ablation study on the dynamic token mechanism. The X-axis represents the number of vision tokens used during inference, while the Y-axis indicates benchmark performance. Three models are trained using a fixed number of vision tokens: 64, 144, and 256. Additionally, a model utilizing the dynamic token mechanism is trained with token counts of $\{64, 144, 256\}$. The $\star$ symbol denotes the number of vision tokens employed during training. The dynamic token mechanism achieved competitive results across various vision token counts.}
    \label{fig:dynamic_token_details}
\end{figure*}

\subsection{Ablation Results} \label{sec:ablation}
In this section, we conducted detailed ablation experiments to validate the effectiveness of the key components of TokenFLEX.

\textbf{Dynamic Token Mechanism.}
We compare the performance of dynamic token training with conventional fixed token training. In the experiments, we adopt a baseline architecture with a basic projector, denoted as $\mathcal{P}_{\text{naive}}$, which incorporates adaptive average pooling followed by 2-layer MLP.
This setup functions as a dynamic token projector, allowing flexible adjustment of image token count.
For dynamic token training, we set the image token numbers to $\Phi = \{64, 144, 256\}$ with proportions of $2:3:5$. In contrast, fixed token training involved training three individual models with 64, 144, and 256 tokens, respectively. The comparison results are presented in Figure~\ref{fig:dynamic_token_details} and Table~\ref{tab:ablation_dynamic_token}.

As shown in Table~\ref{tab:ablation_dynamic_token}, dynamic token training achieves performance competitive with fixed token training across all token configurations. It outperforms fixed models at 64 and 100 tokens, while maintaining comparable scores at 144, 196, and 256 tokens. For instance, dynamic training with 256 tokens scored 60.7\% on OpenCompass, nearly matching to the 61.1\% score of the fixed model. Notably, at 64 tokens, dynamic training scored 58.7\% surpasses the fixed token training score of 57.5\% by 1.2\%. Furthermore, Table~\ref{tab:ablation_dynamic_token} demonstrates robust performance of dynamic training even when inference token counts (e.g., 100 or 196) differ from those used during training.

Another observation from Table~\ref{tab:ablation_dynamic_token} is the lack of performance gains when increasing token counts during inference for fixed-trained models. a model trained with 64 tokens shows minimal improvement as token counts increase (57.5\% at 64 tokens to 56.2\% at 256 tokens). Additionally, models trained with larger token counts (e.g., 256) exhibit performance degradation when inference tokens are reduced (e.g., 61.1\% at 256 tokens drops to 53.3\% at 64 tokens, underperforming the 57.5\% of 64-token fixed model). These findings indicate that fixed-token models cannot adapt to token count variations, whereas dynamic training enables consistent performance across arbitrary token configurations.

Figure~\ref{fig:dynamic_token_details} further illustrates the benchmark performance of dynamic vs. fixed token training. Dynamic training matches or outperforms fixed approaches on most benchmarks. Notably, performance on tasks like MMStar and AI2D plateaus at 144 tokens, underscoring the inefficiency of using uniform token counts across all tasks as adopted in previous VLMs. This validates the adaptive advantage of our proposed TokenFLEX framework.

We also conducted ablation experiments to investigate the impact of token proportions on performance during dynamic token training, as shown in Table~\ref{tab:dynamic_token_proportion}. Results reveal a nuanced relationship between token proportions and model performance. Increasing the proportion of 64-token data failed to improve low-token performance and even degraded overall performance across all token sizes. In contrast, raising the proportion of 256-token data enhanced its own performance while stabilizing results for smaller token counts. We hypothesize that larger tokens encapsulate complex patterns requiring more data for effective learning. Furthermore, the capabilities learned from larger tokens can transfer to fewer tokens. Consequently, with only 20\% of 64-token data can still produce competitive results, demonstrating efficiency in data utilization.

\begin{table}[]
    \centering
    \begin{tabularx}{\columnwidth}{cc|*{3}{>{\centering\arraybackslash}X}}
    \toprule
    \multirow{2}{*}{\#Training Tokens}   & \multirow{2}{*}{Proportion}  & \multicolumn{3}{c}{\#Inference Token} \\ \cmidrule(lr){3-5} 
                                &                              & 64       & 144       & 256       \\ \midrule
    \multirow{3}{*}{64,144,256} & $5:3:2$                        & 58.5      & 59.5      & 59.6     \\
                                & $1:1:1$                        & 58.9 & 60.1 & 60.6 \\
                                & $2:3:5$                        & 58.7 & 60.0 & 60.7 \\ \bottomrule
    \end{tabularx}
    \caption{The impact of the proportion of different token numbers in the training set. Increasing the ratio of samples with smaller token counts fails to enhance performance on low-token instances while degrading model performance on high-token cases, whereas augmenting the proportion of large-token samples universally improves performance across all token ranges.}
    \label{tab:dynamic_token_proportion}
\end{table}

\textbf{Projector Architecture.}
Our goal is to design a projector that supports the dynamic token mechanism, enabling flexible adjustments in the number of vision tokens while adapting to the complexity introduced by this mechanism. 

\textit{Naive Adaptive Average Pooling.}
We first implemented a baseline projector \(\mathcal{P}_{\text{naive}} \) using a two-step approach: (1) grouping vision tokens into grids, followed by (2) average pooling within each group to generate a pooled feature, which is then mapped to the word embedding space via a two-layer MLP with GELU activations. Despite its simplicity,  \(\mathcal{P}_{\text{naive}} \) achieves competitive performance, as shown in Table~\ref{tab:projector_architecture}, with improvements of +0.2\%/+0.3\%/+0.6\% compared to Ola at 64/144/256 tokens. 

\textit{Enhancement via Cross-Attention.}
To address the potential loss of fine-grained details in low-resolution pooled features~\cite{li2024tokenpacker,liu2024oryx}, we then integrated a cross-attention layer between pooled and grouped features. The pooled visual features \(f_{\text{vis}}^p\) act as the query, while the grouped features \(f_{\text{vis}}^g\) serve as the key and value in the cross-attention.  This enhanced architecture $\mathcal{P}_{\text{attn}}$ improves performance by +0.3\%/+0.2\%/+0.5\% at 64/144/256 tokens, respectively. 
This demonstrates the effectiveness of the cross-attention layer. While the gains are smaller than those reported in prior work~\cite{li2024tokenpacker}, this may reflect the mitigating effect of larger-scale training data reducing architecture-dependent performance gaps.

\textit{Comparison with State-of-the-Art Projectors.}
We benchmarked our projectors against  TokenPacker~\cite{li2024tokenpacker} and Ola~\cite{liu2025ola}, both of which support dynamic token mechanisms. As shown in Table~\ref{tab:projector_architecture}, our proposed projector outperforms all baselines across token configurations, achieving gains of +0.7\%/+1.4\%/+1.4\% over TokenPacker at 64/144/256 tokens. This underscores the effectiveness of our architecture while maintaining adaptability to dynamic token counts.

\begin{table}[]
    \centering
    \begin{tabularx}{\columnwidth}{c|*{3}{>{\centering\arraybackslash}X}}
    \toprule
    \multirow{2}{*}{Method} & \multicolumn{3}{c}{\#Inference Token} \\ \cmidrule(lr){2-4} 
                            & 64       & 144       & 256       \\ \midrule
    TokenPacker~\cite{li2024tokenpacker}             & 58.4      & 59.7      & 60.1     \\
    Ola~\cite{liu2025ola}                     & 58.5      & 59.7      & 60.1     \\ \midrule
    $\mathcal{P}_{\text{naive}}$                      & 58.7      & 60.0      & 60.7     \\
    $\mathcal{P}_{\text{attn}}$                      & 59.0      & 60.2      & 61.2     \\
    Ours                      & 59.1      & 61.1      & 61.5  \\ \bottomrule
    \end{tabularx}
    \caption{Ablation on different projectors within the paradigm of dynamic visual token training. We compared our proposed projector against conventional MLPs, attention-based architectures and recent projection designs for adaptive token manipulation. Across all token configuration ranges, the proposed projector consistently outperformed all baseline projectors on the OpenCompass benchmark.}
    \label{tab:projector_architecture}
\end{table}

\section{Conclusion}

We present TokenFLEX, a vision-language framework that overcomes the fixed-token constraint through two key innovations: 1) a stochastic training paradigm enforcing cross-modal alignment across dynamic token counts, and 2) a token-adaptive projector with adaptive pooling and SwiGLU-based feature weighting. Experiments across eight benchmarks demonstrate consistent improvements over fixed-token baselines (1.6\%/1.0\%/0.4\% gains at 64/144/256 tokens) while reducing training time by 13\%. This work establishes the viability of flexible vision token allocation for modern vision-language modeling.
{
    \small
    \bibliographystyle{ieeenat_fullname}
    \bibliography{main}
}

\end{document}